# Nonnegative Matrix Factorization for Semi-supervised Dimensionality Reduction

Youngmin Cho · Lawrence K. Saul

**Abstract** We show how to incorporate information from labeled examples into nonnegative matrix factorization (NMF), a popular unsupervised learning algorithm for dimensionality reduction. In addition to mapping the data into a space of lower dimensionality, our approach aims to preserve the nonnegative components of the data that are important for classification. We identify these components from the support vectors of large-margin classifiers and derive iterative updates to preserve them in a semi-supervised version of NMF. These updates have a simple multiplicative form like their unsupervised counterparts; they are also guaranteed at each iteration to decrease their loss function—a weighted sum of $I$-divergences that captures the trade-off between unsupervised and supervised learning. We evaluate these updates for dimensionality reduction when they are used as a precursor to linear classification. In this role, we find that they yield much better performance than their unsupervised counterparts. We also find one unexpected benefit of the low dimensional representations discovered by our approach: often they yield more accurate classifiers than both ordinary and transductive SVMs trained in the original input space.

**Keywords** Nonnegative matrix factorization · Semi-supervised learning

## 1 Introduction

In many applications of machine learning, high dimensional data must be mapped into a lower dimensional space where it becomes easier to store, manipulate, and

Youngmin Cho
Department of Computer Science and Engineering, University of California, San Diego, La Jolla, CA, USA
E-mail: yoc002@cs.ucsd.edu

Lawrence K. Saul
Department of Computer Science and Engineering, University of California, San Diego, La Jolla, CA, USA
E-mail: saul@cs.ucsd.edu

model. Unsupervised algorithms for dimensionality reduction have the ability to analyze unlabeled examples—a potentially compelling advantage when the number of unlabeled examples exceeds the number of labeled ones. However, unsupervised algorithms also have a corresponding weakness: they do not always preserve the structure in the data that is important for classification.

The above issues highlight the important role of semi-supervised learning algorithms for dimensionality reduction. Such algorithms are designed to analyze data from a mix of labeled and unlabeled examples. From the labeled examples, they aim to preserve those components of the input space that are needed to distinguish different classes of data. From the unlabeled examples, they aim to prune those components of the input space that do not account for much variability. The question, in both theory and in practice, is how to balance these competing criteria (Chapelle et al. 2006; Zhu and Goldberg 2009).

In this paper, we propose a semi-supervised framework for nonnegative matrix factorization (NMF), one of the most popular learning algorithms for dimensionality reduction (Lee and Seung 1999). Widely used for unsupervised learning of text, images, and audio, NMF is especially well-suited to the high-dimensional feature vectors that arise from word-document counts and bag-of-feature representations.

We study the setting where NMF is used as a precursor to large-margin classification by linear support vector machines (SVMs) (Cortes and Vapnik 1995). In this setting, we show how to modify the original updates of NMF to incorporate information from labeled examples. Our approach can be used for problems in binary, multiway, and multilabel classification. Experiments on all these problems reveal significant (and sometimes unexpected) benefits from semi-supervised dimensionality reduction.

The organization of this paper is as follows. In section 2, we introduce our model for semi-supervised NMF and compare it to previous related work (Lee et al. 2010; Liu and Wu 2010). In section 3, we present our experimental results, evaluating a number of competing approaches for semi-supervised learning and dimensionality reduction. Finally in section 4, we review our most significant findings and conclude by discussing directions for future work.

## 2 Model

Nonnegative matrix factorization (NMF) is a method for discovering low dimensional representations of nonnegative data (Lee and Seung 1999). Let $\mathbf{X}$ denote the $d \times n$ matrix formed by adjoining $n$ nonnegative column vectors (*inputs*) of dimensionality $d$. NMF seeks to discover a low-rank approximation

$$\mathbf{X} \approx \mathbf{VH}, \tag{1}$$

where $\mathbf{V}$ and $\mathbf{H}$ are respectively $d \times r$ and $r \times n$ matrices, and typically $r \ll \min(d, n)$. The columns of $\mathbf{V}$ can be viewed as basis vectors; the columns of $\mathbf{H}$, as reconstruction coefficients. Section 2.1 reviews the original formulation of NMF, while section 2.2 describes our extension for semi-supervised learning. Finally, section 2.3 reviews related work.

2.1 NMF for unsupervised learning

The error of the approximation in eq. (1) can be measured in different ways. Here we consider the $I$-divergence (Csiszar 1975):

$$\mathcal{D}(\mathbf{X}, \mathbf{VH}) = \sum_{ij} \left[ X_{ij} \log \frac{X_{ij}}{(VH)_{ij}} - X_{ij} + (VH)_{ij} \right]. \quad (2)$$

The right hand side of eq. (2) is generally positive, vanishing only if the low-rank approximation $\mathbf{VH}$ perfectly reconstructs the matrix $\mathbf{X}$. Other penalties (e.g., sum of squared error) can also be used to derive NMF algorithms (Lee and Seung 2001), but we do not consider them here. The error in eq. (2) can be minimized by iterating the multiplicative updates:

$$V_{ik} \leftarrow V_{ik} \left[ \frac{\sum_j H_{kj} X_{ij}/(VH)_{ij}}{\sum_j H_{kj}} \right], \quad (3)$$

$$H_{kj} \leftarrow H_{kj} \left[ \frac{\sum_i V_{ik} X_{ij}/(VH)_{ij}}{\sum_i V_{ik}} \right]. \quad (4)$$

Simple to implement, these updates are guaranteed to decrease the approximation error at each iteration. They also provably converge to a stationary point—typically, a local minimum—of eq. (2).

Unsupervised NMF can be viewed as mapping the high-dimensional column vectors of $\mathbf{X}$ into the low-dimensional column vectors of $\mathbf{H}$. Often this mapping has interesting properties, yielding sparse distributed representations of the inputs (Lee and Seung 1999). However, this mapping can also lead to a severe distortion of inner products and distances in the low dimensional space. The distortion arises due to the non-orthogonality of the basis vectors in $\mathbf{V}$. In particular, even an exact low-rank factorization $\mathbf{X} = \mathbf{VH}$ does not imply that $\mathbf{X}^\top \mathbf{X} = \mathbf{H}^\top \mathbf{H}$. We can modify the results of NMF in a simple way to obtain a more geometrically faithful mapping. In particular, let

$$\mathbf{Z} = (\mathbf{V}^\top \mathbf{V})^{1/2} \mathbf{H}. \quad (5)$$

The matrix $\mathbf{Z}$ is the same size as $\mathbf{H}$, and its column vectors provide a similarly low dimensional representation of the inputs $\mathbf{X}$ (though not one that is constrained to be nonnegative). Unlike the matrix $\mathbf{H}$, however, we observe that $\mathbf{Z}^\top \mathbf{Z} \approx \mathbf{X}^\top \mathbf{X}$ if $\mathbf{X} \approx \mathbf{VH}$; thus this mapping preserves inner products in addition to reconstructing the data matrix by a low-rank approximation. This property of $\mathbf{Z}$ provides a useful sanity check when NMF is performed as a preprocessing step for classification by linear SVMs (as we consider in the next section). In particular, it ensures that for sufficiently large $r$, SVMs trained on the low-dimensional columns of $\mathbf{Z}$ obtain essentially the same results as SVMs trained on the full $d$-dimensional columns of $\mathbf{X}$.

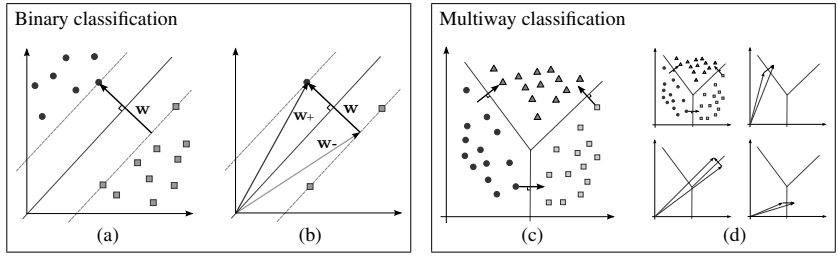

**Fig. 1** *Left*: (a) large-margin hyperplane, with normal vector $\mathbf{w}$, separating positively and negatively labeled examples in the nonnegative orthant; (b) decomposing $\mathbf{w} = \mathbf{w}_+ - \mathbf{w}_-$, where the nonnegative components $\mathbf{w}_+$ and $\mathbf{w}_-$ are derived respectively from support vectors with positive and negative labels. *Right*: (c) important directions in the data for multiway classification, as estimated from the hyperplane decision boundaries between classes; (d) nonnegative components derived from the support vectors in each classifier.

## 2.2 NMF for semi-supervised learning

The updates in eqs. (3–4) are guaranteed to decrease the reconstruction error in eq. (2), but they are not guaranteed to preserve directions in the data that are important for classification. This weakness is not peculiar to NMF: it is the bane of all purely unsupervised algorithms for dimensionality reduction.

### 2.2.1 Binary classification

We begin by addressing this weakness in the simple case of binary classification. Let $\mathbf{x}_i$ denote the $i$th column of the matrix $\mathbf{X}$, and suppose (without loss of generality) that the first $m \leq n$ inputs have binary labels $y_i \in \{-1, +1\}$. For nonnegative data, all these inputs will lie in the same orthant. Fig. 1 (left) shows a hyperplane separating positively and negatively labeled examples in this orthant. The normal vector $\mathbf{w}$ to this hyperplane identifies an important direction for classification. How can we modify NMF so that it preserves this component of the data? This problem is complicated by the fact that the weight vector $\mathbf{w}$ may not itself lie inside the nonnegative orthant. In particular, as shown in the figure, it may have both positive and negative projections onto the orthogonal coordinate axes of the input space. This direction cannot therefore be preserved by any trivial modification of NMF.

We can make progress by appealing to a dual representation of the weight vector $\mathbf{w}$. In such a representation, the vector $\mathbf{w}$ is expressed as a weighted sum of labeled examples,

$$\mathbf{w} = \sum_{i=1}^{m} \alpha_i y_i \mathbf{x}_i, \tag{6}$$

with nonnegative coefficients $\alpha_i \geq 0$. Such a representation arises naturally in large-margin classifiers, where the nonzero coefficients identify support vectors, and in perceptrons, where they count the mistakes made by an online classifier. The dual representation in eq. (6) suggests a simple decomposition of the weight vector $\mathbf{w}$ into nonnegative components. In particular, we define:

$$\mathbf{w}_+ = \sum_{y_i=+1} \alpha_i \mathbf{x}_i, \qquad \mathbf{w}_- = \sum_{y_i=-1} \alpha_i \mathbf{x}_i, \qquad (7)$$

where the sums in eq. (7) range over positively and negatively labeled examples, respectively. Note that both $\mathbf{w}_+$ and $\mathbf{w}_-$ lie in the same orthant as the inputs $\mathbf{x}_i$. In terms of these components, the weight vector is given by the difference:

$$\mathbf{w} = \mathbf{w}_+ - \mathbf{w}_-. \qquad (8)$$

Fig. 1 (b) illustrates this decomposition for a large-margin hyperplane, where the non-zero coefficients $\alpha_i > 0$ identify support vectors.

Though the weight vector $\mathbf{w} \in \Re^d$ may not lie in the same orthant as the inputs, we can modify NMF to preserve the individual components $\mathbf{w}_\pm \in \Re^d_+$ (which are necessarily nonnegative). For ease of notation, we define the nonnegative vectors $\boldsymbol{\alpha}_+ \in \Re^n_+$ and $\boldsymbol{\alpha}_- \in \Re^n_+$ with elements:

$$[\boldsymbol{\alpha}_\pm]_i = \begin{cases} \alpha_i \max(0, \pm y_i) & \text{if } i \leq m, \\ 0 & \text{if } i > m. \end{cases} \qquad (9)$$

In terms of these elements, we can rewrite the nonnegative components in eq. (7) as:

$$\mathbf{w}_+ = \mathbf{X}\boldsymbol{\alpha}_+, \qquad \mathbf{w}_- = \mathbf{X}\boldsymbol{\alpha}_-. \qquad (10)$$

How well are these components preserved by the dimensionality reduction $\mathbf{X} \approx \mathbf{VH}$ in eq. (1) of NMF? Analogous to eq. (10), let $\hat{\mathbf{w}}_+$ and $\hat{\mathbf{w}}_-$ indicate the *reconstructed* components:

$$\hat{\mathbf{w}}_+ = \mathbf{VH}\boldsymbol{\alpha}_+, \qquad \hat{\mathbf{w}}_- = \mathbf{VH}\boldsymbol{\alpha}_-. \qquad (11)$$

The nonnegative components of the weight vector $\mathbf{w}$ will be preserved if both $\mathbf{w}_+ \approx \hat{\mathbf{w}}_+$ and $\mathbf{w}_- \approx \hat{\mathbf{w}}_-$. On the other hand, if these approximations do not hold—if these components are not preserved—then the dimensionality reduction from NMF is likely to shrink the margin of separation between different classes.

Intuitively, we can view the components $\mathbf{w}_\pm$ as additional inputs that must be accurately modeled when NMF is used as a preprocessing step for linear classification. Eq. (10) shows how these additional inputs are constructed from labeled examples in the original input space. Of course, the same labeled examples may not emerge as support vectors when a linear SVM is *retrained* on the low dimensional outputs of NMF. Indeed, the smaller the number of labeled examples, the weaker the hints they provide for semi-supervised dimensionality reduction. The crucial question in this regime is how to best exploit the few labeled examples that are available. In our case, the answer to this question must also take into account that the resulting low dimensional representations will be used as features for linear classification. Our approach—to preserve the direction perpendicular to the large margin hyperplane—is inspired by the statistical learning theory of SVMs. We view the large margin hyperplane as a particularly robust "hint" that can be extracted from a small number of labeled examples in the high dimensional input space.

We can bias NMF to preserve the components $\mathbf{w}_\pm$ by adding extra terms to its loss function. Specifically, for problems in binary classification, we imagine that a weight vector $\mathbf{w}$ has been found by training a perceptron or SVM on $m \leq n$ labeled examples. Then, from the dual representation of $\mathbf{w}$ in eq. (6), we compute the nonnegative components $\mathbf{w}_\pm$ in eq. (7). Finally, to derive a semi-supervised version of NMF, we seek a factorization $\mathbf{X} = \mathbf{VH}$ that minimizes the loss function

$$\mathcal{L} = \mathcal{D}(\mathbf{X},\mathbf{VH}) + \lambda\left[\mathcal{D}(\mathbf{w}_+,\hat{\mathbf{w}}_+) + \mathcal{D}(\mathbf{w}_-,\hat{\mathbf{w}}_-)\right], \qquad (12)$$

where the reconstructed components $\hat{\mathbf{w}}_\pm$ are defined from eq. (11). The first term in eq. (12) is simply the usual $I$-divergence for unsupervised NMF, while the other terms penalize large divergences between the nonnegative components $\mathbf{w}_\pm$ and their reconstructions $\hat{\mathbf{w}}_\pm$. The parameter $\lambda > 0$ determines the balance between unsupervised and supervised learning.

With a slight change in notation, we can rewrite the loss function in eq. (12) in a more symmetric form. In particular, let

$$\mathbf{S} = \begin{bmatrix} \boldsymbol{\alpha}_+ & \boldsymbol{\alpha}_- \end{bmatrix} \qquad (13)$$

denote the $n \times 2$ matrix obtained by adjoining the nonnegative vectors $\boldsymbol{\alpha}_+$ and $\boldsymbol{\alpha}_-$ of support vector coefficients. (The last $n-m$ rows of this matrix are necessarily zero, since unlabeled examples cannot appear as support vectors.) In terms of the matrix $\mathbf{S}$, we can rewrite eq. (12) as:

$$\mathcal{L} = \mathcal{D}(\mathbf{X}, \mathbf{VH}) + \lambda \mathcal{D}(\mathbf{XS}, \mathbf{VHS}). \qquad (14)$$

The goal of semi-supervised NMF is to minimize the right hand side of eq. (14) with respect to the matrices $\mathbf{V}$ and $\mathbf{H}$. Note that we do not update the coefficients $\boldsymbol{\alpha}_\pm$ that appear in the matrix $\mathbf{S}$; these coefficients are estimated prior to NMF by training a linear classifier on the subset of $m$ labeled examples. Once estimated in this way, the matrix $\mathbf{S}$ is assumed to be fixed.

The weighted combination of $I$-divergences in eq. (14) can be minimized by simple multiplicative updates. These updates take the form:

$$V_{ik} \leftarrow V_{ik} \left[ \frac{\sum_j \frac{X_{ij} H_{kj}}{(VH)_{ij}} + \lambda \sum_\ell \frac{(XS)_{i\ell}(HS)_{k\ell}}{(VHS)_{i\ell}}}{\sum_j H_{kj} + \lambda \sum_\ell (HS)_{k\ell}} \right], \qquad (15)$$

$$H_{kj} \leftarrow H_{kj} \left[ \frac{\sum_i \frac{X_{ij} V_{ik}}{(VH)_{ij}} + \lambda \sum_{i\ell} \frac{(XS)_{i\ell} V_{ik} S_{j\ell}}{(VHS)_{i\ell}}}{\sum_i V_{ik} + \lambda \sum_{i\ell} V_{ik} S_{j\ell}} \right]. \qquad (16)$$

The updates require similar matrix computations as eqs. (3–4) and reduce to their unsupervised counterparts when $\lambda = 0$. In the appendix, we show how to derive these updates from an auxiliary function and prove that they monotonically decrease the loss function in eq. (14). We call this model as **NMF-$\boldsymbol{\alpha}$** due to the important role played by the dual coefficients in eq. (6).

*2.2.2 Multiway and multilabel classification*

The framework in the previous section extends in a straightforward way to preserve multiple directions that are important for classification. The need to preserve multiple directions arises in problems of multiway and multilabel classification. For multiway classification, important directions in the data can be estimated by training $c(c-1)/2$ linear classifiers to distinguish all pairs of $c$ classes. Then, just as in the binary case, the normal to each hyperplane decision boundary between classes can be decomposed into a difference of nonnegative components $\mathbf{w}_\pm$. Fig. 1 (d) illustrates the six nonnegative components that arise in this way from a problem with $c=3$ classes. For multilabel classification, we can train as many linear classifiers as there are labels. Then, from each of these classifiers, we can extract nonnegative components $\mathbf{w}_\pm$ to be preserved by NMF.

As in section 2.2.1, we imagine that these important directions are estimated prior to NMF using whatever labeled examples are available for this purpose. Let $\tau \in \{1, 2, ..., p\}$ index the multiple linear SVMs that are trained using these examples. As before, from each SVM, using eqs. (6–9), we can extract the support vector coefficients $\boldsymbol{\alpha}_\pm^{(\tau)}$. Finally, let

$$\mathbf{S} = \begin{bmatrix} \boldsymbol{\alpha}_+^1 \ \boldsymbol{\alpha}_+^2 \ \ldots \ \boldsymbol{\alpha}_+^p \ \boldsymbol{\alpha}_-^1 \ \boldsymbol{\alpha}_-^2 \ \ldots \ \boldsymbol{\alpha}_-^p \end{bmatrix} \qquad (17)$$

denote the $n \times 2p$ matrix that adjoins all the column vectors $\boldsymbol{\alpha}_\pm^{(\tau)}$. Defining the matrix $\mathbf{S}$ in this way, we can perform semi-supervised NMF using the same objective and updates as eqs. (14–16).

2.3 Related work

Several previous authors have modified NMF to incorporate information from labeled examples, though not exactly in the same way. Closest to our approach is the work by Lee et al. (2010), who adapt NMF so that the low dimensional representations $\mathbf{H}$ in eq. (1) are also predictive of the classes of labeled examples. In addition to the nonnegative input matrix $\mathbf{X}$, they construct a binary label matrix $\mathbf{Y}$ whose elements $Y_{kl}$ indicate whether or not the $l$th training example ($l \leq m$) belongs to the $k$th class. Then they attempt to model the relationship between $\mathbf{Y}$ and $\mathbf{H}_{1:m}$ (the first $m$ columns of $\mathbf{H}$) by a linear regression $\mathbf{Y} \approx \mathbf{U}\mathbf{H}_{1:m}$. Combining the sum of squared errors from this regression and the original NMF, they consider the loss function

$$\mathcal{L}(\mathbf{V}, \mathbf{H}, \mathbf{U}) \;=\; \|\mathbf{X} - \mathbf{V}\mathbf{H}\|^2 + \lambda \|\mathbf{Y} - \mathbf{U}\mathbf{H}_{1:m}\|^2, \qquad (18)$$

where the parameter $\lambda > 0$ balances the trade-off between unsupervised and supervised learning, just as in eq. (12). Lee et al. (2010) also show how to minimize eq. (18) using simple multiplicative updates. The model in eq. (18) has obvious parallels to our approach; we therefore evaluate it as an additional baseline in our experiments. (See section 3.)

Another line of related work is the model for constrained nonnegative matrix factorization (CNMF) proposed by Liu and Wu (2010). The main idea in CNMF is to require the data points $\mathbf{x}_i$ and $\mathbf{x}_j$ (i.e., the $i$th and $j$th columns of $\mathbf{X}$) to have the same low-dimensional representation $\mathbf{h}_i = \mathbf{h}_j$ if they belong to the same class. Specifically, $\mathbf{H}$ is first separated into two parts: $\mathbf{H}_{1:m}$ (labeled) and $\mathbf{H}_{m+1:n}$ (unlabeled). Then the reconstruction coefficients for labeled examples are constrained to be generated by the binary label matrix $\mathbf{Y}$: in other words, CNMF requires that $\mathbf{H}_{1:m} = \mathbf{QY}$ for some nonnegative matrix $\mathbf{Q}$. The reconstruction coefficients $\mathbf{H}_{m+1:n}$ for unlabeled examples are not constrained except to be generally nonnegative. Both these conditions can be expressed by writing

$$\mathbf{H} = \mathbf{PA} \quad \text{where} \quad \mathbf{P} = \begin{pmatrix} \mathbf{Q} & \mathbf{H}_{m+1:n} \end{pmatrix} \text{ and } \mathbf{A} = \begin{pmatrix} \mathbf{Y} & 0 \\ 0 & \mathbf{I}_{n-m} \end{pmatrix}. \quad (19)$$

Here the lower-right sub-block $\mathbf{I}_{n-m}$ of the matrix $\mathbf{A}$ denotes an identity matrix of size $(n-m) \times (n-m)$. By plugging $\mathbf{H}$ in eq. (19) into the sum of squared errors of the original NMF, they obtain the loss function:

$$\mathcal{L}(\mathbf{V}, \mathbf{P}) = \|\mathbf{X} - \mathbf{VPA}\|, \quad (20)$$

which is minimized by simple multiplicative updates for $\mathbf{V}$ and $\mathbf{P}$. Note that the loss function in eq. (20) does not involve a tuning parameter, such as the parameter $\lambda$ that appears in eqs. (12) and (18). We also evaluate the model in eq. (20) as an additional baseline in our experiments.

In addition to these two baselines, we briefly mention other related approaches. Wang et al. (2004) and Heiler and Schnörr (2006) modified NMF based on ideas from linear discriminant analysis, adapting the reconstruction coefficients in $\mathbf{H}$ to optimize intra-class and inter-class variances. Chen et al. (2007) proposed a nonnegative tri-factorization of Gram matrices for clustering with side information. Wang et al. (2009) proposed a semi-supervised version of NMF based on ideas from graph embedding, representing the similarity between examples (both labeled and unlabeled) by the edges of a weighted graph. In comparison to the work by Lee et al. (2010) and Liu and Wu (2010), we found these other methods to be either less competitive or considerably more difficult to implement, involving many tuning parameters and/or more expensive optimizations. Thus in this paper we do not include their results.

Note that we are interested in semi-supervised NMF for extremely large data sets of sparse, high dimensional inputs (e.g., word-document counts). We do not expect such inputs to be densely sampled from a low dimensional manifold, a common assumption behind many graph-based semi-supervised algorithms. Our own approach exploits the dual representation of linear SVMs and perceptrons because these algorithms scale very well to massively large data sets. These considerations have also informed our survey of previous work.

## 3 Experiments

We experimented with our approach on three publicly available data sets. For each data set, we separated the examples into training, validation, and test sets, as shown in Table 1. We evaluated the benefits of semi-supervised learning by varying the number of training examples whose labels were available for dimensionality reduction and large-margin classification. For each experiment, we averaged our results over five different splits of training and validation examples. However, we used the same test sets across all experiments; this was done to be consistent with previous benchmarks, which specifically designated the examples that appeared in the test set.

**Table 1** Composition of data sets for binary, multiway, and multilabel classification.

|                        | MNIST | News  | Aviation |
|-----------------------:|------:|------:|---------:|
| Examples (training)    | 5000  | 9007  | 12912    |
| Examples (validation)  | 6791  | 2262  | 8607     |
| Examples (test)        | 1991  | 7505  | 7077     |
| Input dimensionality   | 784   | 53975 | 28426    |
| Classes or labels      | 2     | 20    | 22       |

We chose the data sets in Table 1 mainly to experiment on three different types of tasks. The first task—in *binary classification*—was to distinguish grayscale images of 4s versus 9s from the MNIST data set of handwritten digits (LeCun and Cortes 1998). The second task—in *multiway classification*—was to label postings from the 20-Newsgroups[1] data set. Finally, the third task—in *multilabel classification*—was to recognize aviation safety reports[2] from 22 non-exclusive categories. For the first two tasks, we measured performance by classification accuracy; for the third task, in multilabel classification, we computed the average of macro-averaged and micro-averaged F-measures.

The News and Aviation data sets in Table 1 typify the sorts of applications we imagine for semi-supervised NMF: these data sets consist of sparse, high dimensional inputs, of the sort that naturally arise from word-document counts and bag-of-feature representations. The MNIST data set does not have this structure, but as we shall see, it is useful for visualizing the effect of label information on the basis vectors discovered by NMF.

3.1 Methodology

Each of our experiments was designed to measure the performance when dimensionality reduction of some kind is followed by large-margin classification. We compared six methods for dimensionality reduction: principal component analysis (PCA), linear discriminant analysis (LDA), unsupervised NMF with the $I$-divergence in eq. (2), semi-supervised NMF (Lee et al. 2010) and constrained NMF (Liu and Wu 2010) with the loss functions (respectively) in eqs. (18) and (20), and our own approach NMF-$\alpha$. All these methods except LDA[3] operated on the entire set of $n$ training examples, but only the semi-supervised approaches made special use of the $m \leq n$ labeled examples. For NMF-$\alpha$, we began by training one or more linear SVMs on the $m$ labeled training examples, as described in section 2.2. The hyperplane decision boundaries in these SVMs were then fed to the semi-supervised updates in eqs. (15–16). Finally, we used the results of PCA, LDA, and NMF to map all the inputs into a space of dimensionality $r \leq d$. (Note that once the basis vectors from these methods are known, no labels are required to compute this mapping on test examples.) For NMF, we also performed the corrective mapping described at the end of section 2.1.

---

[1] http://www.ai.mit.edu/people/jrennie/20Newsgroups/
[2] http://web.eecs.utk.edu/events/tmw07/
[3] Only the $m$ labeled examples were used in LDA.

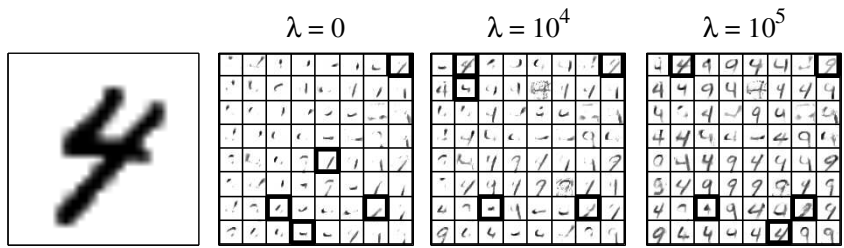

**Fig. 2** Comparison of $r = 64$ basis vectors for images of handwritten digits (4s and 9s) learned by unsupervised and semi-supervised NMF. As $\lambda$ increases from left to right, the basis vectors evolve from primitive strokes to confusable examples. The image on the left is reconstructed by the highlighted basis vectors in each panel.

Next we measured the effects of dimensionality reduction on large-margin classification. Using *only* the same $m$ labeled examples in the training set, we evaluated the performance of linear SVMs on the reduced-dimensional test examples from PCA, LDA, and NMF. We used the labeled examples in the validation set to tune the margin penalty parameter in SVMs as well as the trade-off parameter $\lambda$ in eqs. (14) and (18). For reference, we also tested linear SVMs and transductive SVMs (Joachims 1999b) on the raw inputs (i.e., without any dimensionality reduction). The transductive SVMs made use of all $n$ examples in the training set, both labeled and unlabeled. We used the LIBLINEAR package (Fan et al. 2008) to train the baseline SVMs in these experiments and SVM[light] (Joachims 1999a) to train the transductive SVMs.

For each of the data sets in Table 1, we experimented with different numbers of basis vectors $r \leq d$ for dimensionality reduction and different numbers of labeled training examples $m \leq n$ for semi-supervised learning. The following sections present a representative subset of our results.

3.2 Qualitative results

We begin by illustrating qualitatively how the results from NMF change when the algorithm is modified to incorporate information from labeled examples. Fig. 2 compares the $r = 64$ basis vectors of MNIST handwritten digits (4s and 9s) learned for different values of $\lambda$ in eq. (14). The visual differences in these experiments are striking even though only 2% of the training examples were labeled and the corresponding basis vectors in different panels were initialized in the same way. Note how for $\lambda = 0$ (i.e., purely unsupervised NMF), the basis vectors resemble cursive strokes (Lee and Seung 1999), whereas for very large $\lambda$, they resemble actual training examples (e.g., support vectors). The middle panel shows the results from the intermediate value of $\lambda$ whose basis vectors yielded the best-performing classifier.

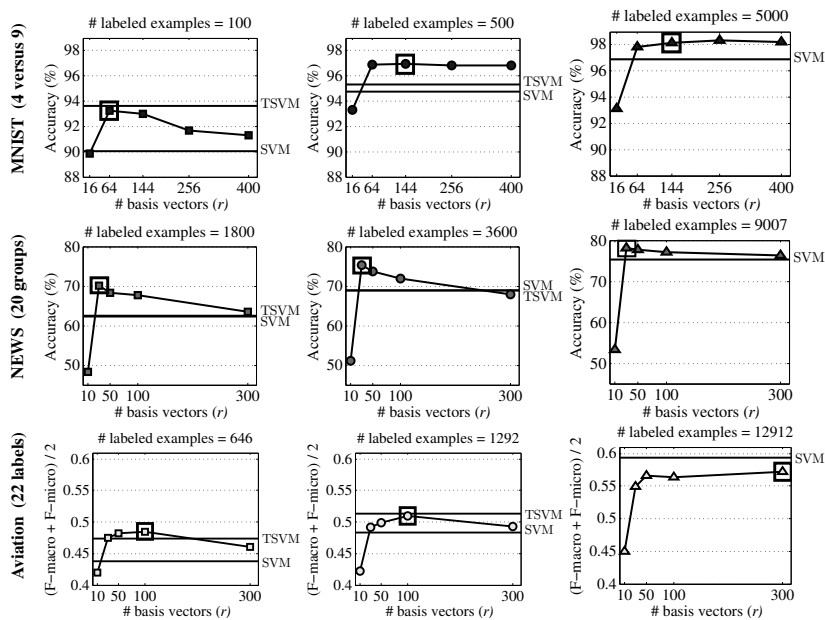

**Fig. 3** Effect of dimensionality reduction by NMF-$\alpha$ on the performance of linear SVMs. Also shown as baselines are the performance of ordinary and transductive SVMs on the original (raw) inputs. (The latter is indicated by TSVM.) Each sub-figure plots the classification accuracy or F-measure on the test set versus the numbers of basis vectors used for dimensionality reduction. The hollow black square indicates the model that performed best on the validation set. *Top row*: Binary classification on the MNIST data set (4s vs. 9s) when 2% (left), 10% (middle), and 100% (right) of training examples were labeled. *Middle row*: Multiway classification on the 20-Newsgroups data set when 20% (left), 40% (middle), and 100% (right) of training examples were labeled. *Bottom row*: Multilabel classification on the Aviation data set when 5% (left), 10% (middle), and 100% (right) of training examples were labeled.

### 3.3 Benefits of dimensionality reduction

Next we examine the interplay between classification performance, semi-supervised learning, and dimensionality reduction. To begin, we compare NMF-$\alpha$ to two other approaches that do *not* involve dimensionality reduction. Fig. 3 shows the results from NMF-$\alpha$ from experiments in binary classification, multiway classification, and multilabel classification. We report the performance in terms of classification accuracy or F-measure on the test examples as we vary both the number of labeled training examples (along different plots in the same row) and the number of basis vectors for dimensionality reduction (along the horizontal axes within each plot). Also shown as baselines are the performance of ordinary and transductive SVMs on the original (raw) inputs.

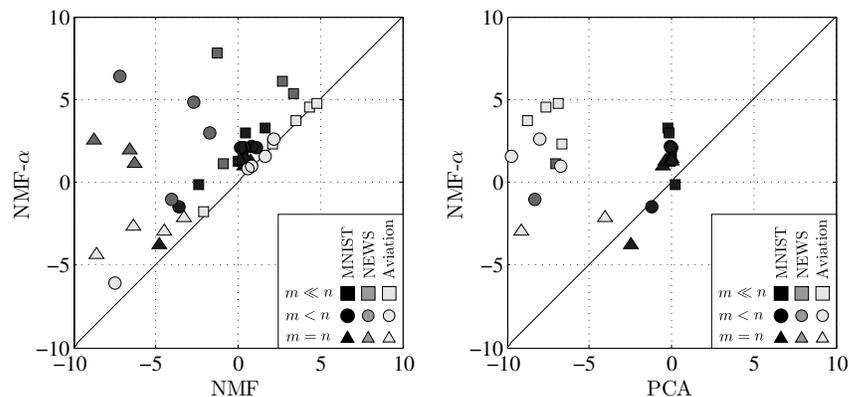

**Fig. 4** Comparing the effects on linear classification of dimensionality reduction by NMF-$\alpha$ versus purely unsupervised methods. The plots compare the results from NMF-$\alpha$ in Fig. 3 to analogous results from ordinary NMF (*left*) and PCA (*right*); each point represents a pair of experiments on the same data set (indicated by color), for the same number of labeled examples (indicated by shape), and with the same number of basis vectors (not indicated in the plot). The axes delineate the classification accuracy (%) or 100×F-measure relative to the baseline result obtained by linear SVMs in the original input space. Thus points in the upper half-space reveal cases where dimensionality reduction by NMF-$\alpha$ improved classification over the baseline SVMs; points in the right half-space reveal cases where dimensionality reduction by NMF or PCA improved classification over the baseline SVMs; finally, points above the diagonal line reveal cases where the dimensionality reduction by NMF-$\alpha$ yielded better performance than purely unsupervised methods. For better visualization, we omit some points where NMF-$\alpha$ performs much better than unsupervised approaches, but the opposite cases are all shown.

The results in Fig. 3 reveal an interesting trend: *the representations from NMF-$\alpha$ often yield better performance than SVMs (and even transductive SVMs) trained on the raw inputs*. In particular, note how in many plots, one or more results from NMF-$\alpha$ top the performance of the baseline and transductive SVM in the original input space. This trend[4] suggests that the updates for NMF-$\alpha$ are not only preserving discriminative directions in the input space, but also pruning away noisy, irrelevant subspaces that would otherwise compromise the performance of large-margin classification.

### 3.4 Benefits of semi-supervised learning

Next we compare the effects on linear classification by different methods of dimensionality reduction. Fig. 4 compares NMF-$\alpha$ in this role to the purely unsupervised methods of NMF and PCA. These latter methods derive basis vectors from all the training examples, but they do not exploit the extra information in the $m$ labeled examples. To produce Fig. 4, we collected the results for ordinary NMF and PCA analogous to those in Fig. 3 for NMF-$\alpha$ ; we then plotted them against the results for NMF-$\alpha$ . At a high level, points above the diagonal line indicate comparable models (i.e., those with the same number of basis vectors) where NMF-$\alpha$ outperforms either NMF (*left*) and PCA (*right*) when these methods are used as a precursor for linear classification. (See the caption for more detail.) Note that the vast majority of points lie above the diagonal line, often by a significant margin. Thus this figure shows clearly the benefits of *semi-supervised* learning for dimensionality reduction.

---

[4] Similar results were observed on the validation sets, suggesting cross-validation as one way to select the number of basis vectors. The hollow black square in each plot indicates the best classifier on the validation set.

**Table 2** Comparing the effects on linear classification of dimensionality reduction by NMF-$\alpha$ versus the purely supervised method of LDA. See text for details.

|  | MNIST | | 20 Newsgroups | | Aviation | |
|---|---|---|---|---|---|---|
|  | NMF-$\alpha$ | LDA | NMF-$\alpha$ | LDA | NMF-$\alpha$ | LDA |
| $m \ll n$ | **93.3** | 88.5 | **70.1** | 65.3 | **48.5** | 43.6 |
| $m < n$ | **96.9** | 92.7 | **75.3** | 69.8 | **51.0** | 47.9 |
| $m = n$ | **98.2** | 94.2 | **78.1** | 73.5 | **57.2** | 56.4 |

We also compared NMF-$\alpha$ to another canonical method for dimensionality reduction: linear discriminant analysis (LDA). LDA can be regarded as a purely *supervised* method that computes informative directions in the data from labeled examples (but not unlabeled ones). LDA is typically used to extract a number of basis vectors equal to one less than the number of classes. Table 2 compares the results in linear classification by SVMs when LDA and NMF-$\alpha$ were used for dimensionality reduction. The results for NMF-$\alpha$ in this table are from the models that obtained the best average performance on the validation set—i.e., those corresponding to the hollow black squares in Fig. 3. Note that LDA is limited by the small number of basis vectors that it can extract. Again these results reveal the benefits of semi-supervised learning for dimensionality reduction: NMF-$\alpha$ outperforms LDA in every experimental setting.

3.5 Comparison to other semi-supervised approaches

Finally we compare the effects on linear classification by dimensionality reduction by NMF-$\alpha$ to other semi-supervised versions of NMF. Fig. 5 compares NMF-$\alpha$ in this role to the semi-supervised methods described in section 2.3. To produce Fig. 4, we evaluated the methods of Lee et al. (2010) and Liu and Wu (2010) with the same numbers of basis vectors and labeled examples as shown in Fig. 3; we then plotted the results for these methods against the results for NMF-$\alpha$ . One clear trend from these plots is that NMF-$\alpha$ works better in the regime $m \ll n$ where there are very few labeled examples. (Note how nearly all the squares lie above the diagonal line.) The plots validate our intuition—based on results in purely supervised learning—that margin-based approaches are better suited to this regime than linear regression. From the plots, we conclude that NMF-$\alpha$ is generally exploiting more robust information in the labeled examples—and this is particularly true when there are very few of them.

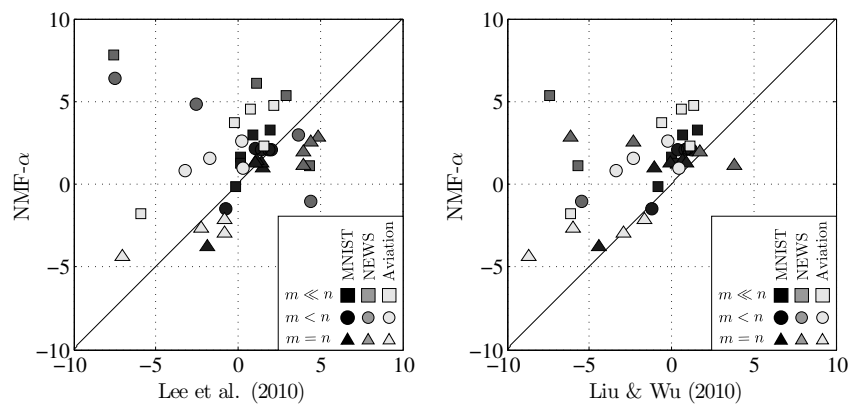

**Fig. 5** Comparing the effects on linear classification of dimensionality reduction by NMF-$\alpha$ versus other semi-supervised versions of NMF: Lee et al. (2010) (*left*) and Liu & Wu (2010) (*right*). The format is the same as Fig. 4.

## 4 Discussion

In this paper we have investigated a novel form of NMF for semi-supervised dimensionality reduction. The updates for our approach are simple to implement and guaranteed to converge. Empirically, we have observed several benefits when these updates are used as a precursor to linear classification. First, as shown in Fig. 3, the lower dimensional representations discovered by NMF-$\alpha$ often lead to better performance than SVMs and transductive SVMs trained in the original input space; this improvement was observed in both semi-supervised and fully supervised settings. Second, as shown in Fig. 4 and Table 2, NMF-$\alpha$ performs better in this role than other canonical methods for dimensionality reduction, such as PCA, ordinary NMF, and LDA. Third, as shown in Fig. 5, NMF-$\alpha$ seems especially well suited to the regime where the number of unlabeled examples greatly exceeds the number of labeled ones. In particular, NMF-$\alpha$ appears to perform better in this regime than other semi-supervised versions of NMF, presumably because its margin-based approach is more robust than least-squares methods. All these results suggest a useful role for NMF-$\alpha$ in problems of high dimensional classification—especially those problems where examples are plentiful but labels are scarce.

While this paper focuses on semi-supervised NMF, many other authors have explored how to incorporate information from labeled examples in related models of matrix factorization. Blei and McAuliffe (2008) proposed a variant of latent Dirichlet allocation for supervised topic modeling of word-document matrices. Rish et al. (2008) used shared basis components from exponential family PCA to build a joint model over inputs and labels, whereas Goldberg et al. (2010) considered directly completing a joint matrix of inputs and labels while minimizing its rank. Mairal et al. (2009) showed how to adapt sparse coding for supervised dictionary learning: the dictionary entries in this scheme are optimized for classification as well as reconstruction accuracy. Our work has similar goals, but by starting in a different place, we are able to capitalize on the many desirable properties (e.g., ease of implementation) of NMF.

Motivated by our results in this area, as well as those of others, we are investigating several directions for future work. For very large-scale problems, where training one or more linear SVMs is not feasible, we are exploring faster online algorithms — perceptrons and recent variants thereof (Dredze et al. 2008) — to estimate which directions in the data should be preserved for classification. We are also considering to jointly optimize the NMF loss functions and the support vector coefficients $\alpha$ for better modeling (Gupta and Xiao 2011). We note that the computations in eqs. (15–16) are easily parallelized; in future work, we hope to exploit this parallelizable structure with GPU programming. Others have noted that NMF can be kernelized (Zhang et al. 2006); we are looking into a kernelized version of our algorithm whose low-rank factorizations support more scalable implementations of nonlinear SVMs (Fine and Scheinberg 2001). Finally, we are interested in applying the ideas in this paper to slightly different settings, such as domain adaptation and transfer learning (Ando and Zhang 2005; Blitzer et al. 2007), where hybrid schemes for dimensionality reduction have proven extremely useful. These directions and others are left for future work.

## A Derivation of multiplicative updates

In this appendix we show how to derive the multiplicative updates in eqs. (15–16). We also sketch the proof that they converge to a stationary point of the loss function in eq. (14). The full proof is based on constructing an auxiliary function that provides an upper bound on the loss function. Our construction closely mimics the one used by Lee and Seung (2001) in their study of NMF for unsupervised learning. Accordingly, we omit many details where the two proofs coincide.

We begin by bounding the individual $I$-divergences that appear in eq. (14). From Jensen's inequality, we know that for any nonnegative numbers $z_k$:

$$\log \sum_k z_k \geq \sum_k \theta_k \log \frac{z_k}{\theta_k}, \tag{21}$$

where $\theta_k > 0$ and $\sum_k \theta_k = 1$. To bound the $I$-divergences in eq. (14), we apply this inequality to the implied sums in the matrix products **VH** and **VHS**. Specifically, we can write:

$$\log(VH)_{ij} \geq \sum_k \eta_{ijk} \log \frac{V_{ik} H_{kj}}{\eta_{ijk}}, \tag{22}$$

$$\log(VHS)_{i\ell} \geq \sum_{jk} \psi_{i\ell jk} \log \frac{V_{ik} H_{kj} S_{j\ell}}{\psi_{i\ell jk}}, \tag{23}$$

where $\eta_{ijk}$ and $\psi_{i\ell jk}$ are positive numbers (chosen later, to achieve the tightest possible bounds) that satisfy $\sum_k \eta_{ijk} = \sum_{jk} \psi_{i\ell jk} = 1$. Substituting these inequalities into eq. (14), we obtain the upper bound:

$$\mathcal{D}(\mathbf{X}, \mathbf{VH}) + \lambda \mathcal{D}(\mathbf{XS}, \mathbf{VHS}) \leq$$
$$\sum_{ijk} X_{ij}\eta_{ijk} \log \frac{X_{ij}\eta_{ijk}}{V_{ik}H_{kj}} + \sum_{ij} \left[(VH)_{ij} - X_{ij}\right] +$$
$$\lambda \sum_{i\ell jk} (XS)_{i\ell}\psi_{i\ell jk} \log \frac{(XS)_{i\ell}\psi_{i\ell jk}}{V_{ik}H_{kj}S_{j\ell}} + \lambda \sum_{i\ell} \left[(VHS)_{i\ell} - (XS)_{i\ell}\right]. \quad (24)$$

The iterative updates in eqs. (15–16) are derived by minimizing the upper bound in eq. (24) in lieu of the original loss function. By setting its partial derivatives to zero, the bound can be minimized in closed form with respect to either of the nonnegative matrices $\mathbf{V}$ or $\mathbf{H}$. With respect to $\mathbf{V}$, the minimum is given by:

$$V_{ik} = \frac{\sum_j X_{ij}\eta_{ijk} + \lambda \sum_{j\ell}(XS)_{i\ell}\psi_{i\ell jk}}{\sum_j H_{kj} + \lambda \sum_\ell (HS)_{k\ell}}. \quad (25)$$

The question remains how to set $\eta_{ijk}$ and $\psi_{i\ell jk}$ in this expression. In brief, it can be shown that by setting them appropriately, we can ensure that the update in eq. (25) monotonically decreases the original loss function in eq. (14). Specifically, the following choices — in terms of the current parameter estimates for $\mathbf{V}$ and $\mathbf{H}$ — lead to this property:

$$\eta_{ijk} = V_{ik}H_{kj}/(VH)_{ij}, \quad (26)$$
$$\psi_{i\ell jk} = V_{ik}H_{kj}S_{j\ell}/(VHS)_{i\ell}. \quad (27)$$

Substituting these values into eq. (25) yields the update rule for $\mathbf{V}$ in eq. (15). The update rule for $\mathbf{H}$ can be derived in an analogous fashion.

**References**

Ando, R. K., & Zhang, T. (2005). A framework for learning predictive structures from multiple tasks and unlabeled data. *Journal of Machine Learning Research*, *6*, 1817–1853.

Blei, D., & McAuliffe, J. (2008). Supervised topic models. In Platt, J. C., Koller, D., Singer, Y., & Roweis, S. (Eds.), *Advances in Neural Information Processing Systems 20* (pp. 121–128). Cambridge, MA: MIT Press.

Blitzer, J., Dredze, M., & Pereira, F. (2007). Biographies, Bollywood, boom-boxes and blenders: domain adaptation for sentiment classification. In *Proceedings of the 45th Annual Meeting of the Association of Computational Linguistics (ACL-07)* (pp. 440–447). Prague, Czech Republic.

Chapelle, O., Schölkopf, B., & Zien, A. (Eds.). (2006). *Semi-Supervised Learning*. MIT Press.

Chen, Y., Rege, M., Dong, M., & Hua, J. (2007). Incorporating user provided constraints into document clustering. In *Proceedings of the Seventh IEEE International Conference on Data Mining (ICDM-07)* (pp. 103–112). Washington, DC, USA.

Cortes, C., & Vapnik, V. (1995). Support-vector networks. *Machine Learning*, *20*, 273–297.

Csiszar, I. (1975). *I*-divergence geometry of probability distributions and minimization problems. *The Annals of Probability*, *3*(1), 146–158.

Dredze, M., Crammer, K., & Pereira, F. (2008). Confidence-weighted linear classification. In *Proceedings of the 25th International Conference on Machine Learning (ICML-08)* (pp. 264–271). Helsinki, Finland.

Fan, R.-E., Chang, K.-W., Hsieh, C.-J., Wang, X.-R., & Lin, C.-J. (2008). LIBLINEAR: A library for large linear classification. *Journal of Machine Learning Research*, *9*, 1871–1874.

Fine, S., & Scheinberg, K. (2001). Efficient SVM training using low-rank kernel representations. *Journal of Machine Learning Research*, *2*, 243–264.

Goldberg, A., Zhu, X., Recht, B., Xu, J., & Nowak, R. (2010). Transduction with matrix completion: Three birds with one stone. In Lafferty, J., Williams, C. K. I., Shawe-Taylor, J., Zemel, R., & Culotta, A. (Eds.), *Advances in Neural Information Processing Systems 23* (pp. 757–765).

Gupta, M. D., & Xiao, J. (2011). Non-negative matrix factorization as a feature selection tool for maximum margin classifiers. In *Proceedings of the Twenty-Fourth IEEE Conference on Computer Vision and Pattern Recognition (CVPR-11)* (pp. 2841–2848).

Heiler, M., & Schnörr, C. (2006). Learning sparse representations by non-negative matrix factorization and sequential cone programming. *Journal of Machine Learning Research*, *7*, 1385–1407.

Joachims, T. (1999a). Making large-scale SVM learning practical. In B. Schölkopf, C. Burges and A. Smola (Eds.), *Advances in Kernel Methods - Support Vector Learning* chapter 11, (pp. 169–184). Cambridge, MA: MIT Press.

Joachims, T. (1999b). Transductive inference for text classification using support vector machines. In *Proceedings of the 16th International Conference on Machine Learning (ICML-99)* (pp. 200–209). Bled, Slovenia.

LeCun, Y., & Cortes, C. (1998). The MNIST database of handwritten digits. http://yann.lecun.com/exdb/mnist/.

Lee, D. D., & Seung, H. S. (1999). Learning the parts of objects by non-negative matrix factorization. *Nature*, *401*(6755), 788–791.

Lee, D. D., & Seung, H. S. (2001). Algorithms for non-negative matrix factorization. In Leen, T. K., Dietterich, T. G., & Tresp, V. (Eds.), *Advances in Neural Information Processing Systems 13* (pp. 556–562). Cambridge, MA: MIT Press.

Lee, H., Yoo, J., & Choi, S. (2010). Semi-supervised nonnegative matrix factorization. *IEEE Signal Processing Letters*, *17*(1), 4–7.

Liu, H., & Wu, Z. (2010). Non-negative Matrix Factorization with Constraints. In *Proceedings of the Twenty-Fourth AAAI Conference on Artificial Intelligence (AAAI-10)* (pp. 506–511).

Mairal, J., Bach, F., Ponce, J., Sapiro, G., & Zisserman, A. (2009). Supervised dictionary learning. In Koller, D., Schuurmans, D., Bengio, Y., & Bottou, L. (Eds.), *Advances in Neural Information Processing Systems 21* (pp. 1033–1040).

Rish, I., Grabarnik, G., Cecchi, G., Pereira, F., & Gordon, G. J. (2008). Closed-form supervised dimensionality reduction with generalized linear models. In *Proceedings of the 25th International Conference on Machine Learning (ICML-08)* (pp. 832–839). Helsinki, Finland.

Wang, C., Yan, S., Zhang, L., & Zhang, H. (2009). Non-negative semi-supervised learning. In *Proceedings of the 12th International Conference on Artificial Intelligence and Statistics*.

Wang, Y., Jia, Y., Hu, C., & Turk, M. (2004). Fisher non-negative matrix factorization for learning local features. In *Asian Conference on Computer Vision* (pp. 806–811).

Zhang, D., Zhou, Z.-H., & Chen, S. (2006). Non-negative matrix factorization on kernels. In *PRICAI 2006: Trends in Artificial Intelligence* (pp. 404–412).

Zhu, X., & Goldberg, A. B. (2009). *Introduction to Semi-Supervised Learning*. Morgan & Claypool Publishers.